\documentclass[nohyperref]{article}

\usepackage{hyperref}
\usepackage[accepted]{icml2022}

\usepackage{microtype}
\usepackage{graphicx}
\usepackage{subfigure}
\usepackage{booktabs} %
\usepackage{ragged2e}

  \definecolor{mydarkblue}{rgb}{0,0.08,0.45}
  \hypersetup{ %
    pdftitle={},
    pdfauthor={},
    pdfsubject={Proceedings of the International Conference on Machine Learning 2022},
    pdfkeywords={},
    pdfborder=0 0 0,
    pdfpagemode=UseNone,
    colorlinks=true,
    linkcolor=mydarkblue,
    citecolor=mydarkblue,
    filecolor=mydarkblue,
    urlcolor=mydarkblue,
    }
\usepackage{algorithm}
\usepackage{algorithmic}

\usepackage{amsmath}
\usepackage{amssymb}
\usepackage{mathtools}
\usepackage{amsthm}
\usepackage{multirow}

\newcommand{\calP}{\mathcal{P}}

\newcommand{\calX}{\mathcal{X}}
\newcommand{\calY}{\mathcal{Y}}

\newcommand{\bx}{\mathbf{x}}
\newcommand{\bz}{\mathbf{z}}

\newcommand{\method}{{KNN}}
\newcommand{\methodplus}{{KNN+}}

\usepackage[capitalize,noabbrev]{cleveref}

\theoremstyle{plain}
\newtheorem{theorem}{Theorem}[section]
\newtheorem*{theorem*}{Theorem}

\newtheorem{lemma}[theorem]{Lemma}

\theoremstyle{definition}

\theoremstyle{remark}

\usepackage[textsize=tiny]{todonotes}

\begin{document}

\twocolumn[
\icmltitle{Out-of-Distribution Detection with Deep Nearest Neighbors}

\begin{icmlauthorlist}
\icmlauthor{Yiyou Sun}{wisc}
\icmlauthor{Yifei Ming}{wisc}
\icmlauthor{Xiaojin Zhu}{wisc}
\icmlauthor{Yixuan Li}{wisc}
\end{icmlauthorlist}

\icmlaffiliation{wisc}{Computer Sciences Department, University of Wisconsin-Madison}

\icmlcorrespondingauthor{Yiyou Sun, Yixuan Li}{{sunyiyou, sharonli@cs.wisc.edu}}

\icmlaffiliation{wisc}{Department of Computer Sciences, University of Wisconsin - Madison}
\icmlkeywords{Out-of-distribution Detection}

\vskip 0.3in
]

\printAffiliationsAndNotice{}  %

\begin{abstract}
Out-of-distribution (OOD) detection is a critical task for deploying machine learning models in the open world. Distance-based methods have demonstrated promise, where testing samples are detected as OOD if they are relatively far away from in-distribution (ID) data. However, prior methods impose a strong distributional assumption of the underlying feature space, which may not always hold. In this paper, we explore the efficacy of non-parametric nearest-neighbor distance for OOD detection, which has been largely overlooked in the literature. Unlike prior works, our method does not impose any distributional assumption, hence providing stronger flexibility and generality. We demonstrate the effectiveness of nearest-neighbor-based OOD detection on several benchmarks and establish superior performance. Under the same model trained on ImageNet-1k, our method substantially reduces the false positive rate (FPR@TPR95) by {24.77}\% compared to a strong baseline SSD+, which uses a parametric approach Mahalanobis distance in detection. {Code is available: \url{https://github.com/deeplearning-wisc/knn-ood}}.
\end{abstract}

%%%%%%%%%%%%%%%%%%%%%%%%%%%%%%%%%%%%%%%%%%%%%%%%%%%%%%%%%%%%%%
%%%%%%%%%%%%%%%%%%%%  INTRODUCTION SECTION %%%%%%%%%%%%%%%%%%%
%%%%%%%%%%%%%%%%%%%%%%%%%%%%%%%%%%%%%%%%%%%%%%%%%%%%%%%%%%%%%%
\section{Introduction}
\label{sec:intro}

Modern machine learning models deployed in the open world often struggle with out-of-distribution (OOD) inputs---samples from a different distribution that the network has not been exposed to during training, and therefore should not be predicted at test time. A reliable classifier should not only accurately classify known in-distribution (ID) samples, but also identify as ``unknown'' any OOD input. This gives rise to the importance of OOD detection, which determines whether an input is ID or OOD and enables the
model to take precautions.

A rich line of OOD detection algorithms has been developed recently, among which distance-based methods demonstrated promise~\citep{lee2018simple, tack2020csi, 2021ssd}. Distance-based methods leverage feature embeddings extracted from a model, and operate under the assumption that the test OOD samples are relatively far away from the ID data. For example, \citeauthor{lee2018simple} modeled the feature embedding space as a mixture of multivariate Gaussian distributions, and used the maximum Mahalanobis distance~\citep{mahalanobis1936generalized} to all class centroids for OOD detection. 
However, all these approaches make a strong distributional assumption of the underlying feature space being class-conditional Gaussian. 
As we verify, the learned embeddings can fail the Henze-Zirkler multivariate normality test~\citep{henze1990class}. This limitation leads to the open question:

\begin{center}
\emph{Can we leverage the non-parametric nearest neighbor approach for OOD detection?}
\end{center}

Unlike prior works, the non-parametric approach does not impose {any} distributional assumption about the underlying feature space, hence providing stronger \emph{flexibility and generality}. Despite its simplicity, the nearest neighbor approach has received scant attention. Looking at the literature on OOD detection in the past several years, there has not been any work that demonstrated the efficacy of a non-parametric nearest neighbor approach for this problem. {This suggests that making the seemingly simple idea work is non-trivial}. Indeed, we found that simply using the nearest neighbor distance derived from the feature embedding of a standard classification model is not performant.

\begin{figure*}[t]
\centering
\includegraphics[width=0.99\linewidth]{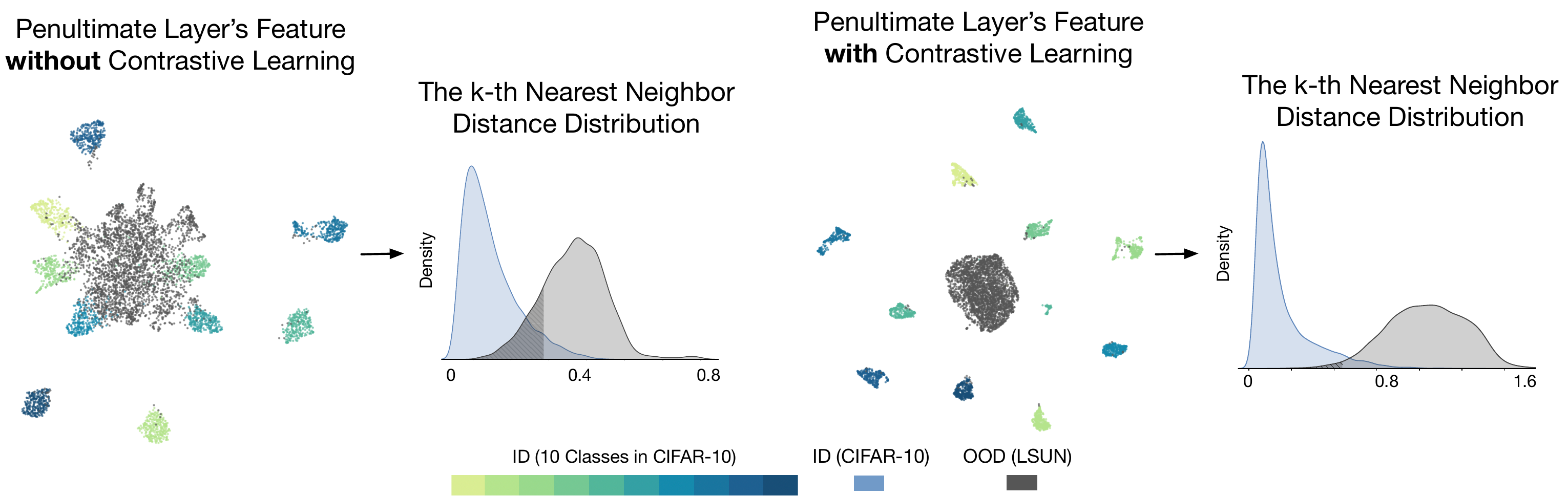}
\caption{\small Illustration of our framework using nearest neighbors for OOD detection. \method performs non-parametric level set estimation, partitioning the data into two sets (ID vs. OOD) based on the $k$-th nearest neighbor distance. The distances are estimated from the penultimate feature embeddings, visualized via UMAP~\cite{mcinnes2018umap-software}.
Models are trained on ResNet-18~\cite{he2016identity} using cross-entropy loss (left) v.s. contrastive loss (right). The in-distribution data is CIFAR-10 (colored in non-gray colors) and  OOD data is LSUN (colored in gray). The shaded grey area in the density distribution plot indicates OOD samples that are misidentified as ID data. }
\label{fig:umap}
\end{figure*}

In this paper, {we challenge the status quo by presenting the first study exploring and demonstrating the efficacy of the non-parametric nearest-neighbor distance for OOD detection}. To detect OOD samples, we compute the  $k$-th nearest neighbor (KNN) distance between the embedding of test input and the embeddings of the training set and use
a threshold-based criterion to determine if the input
is OOD or not. In a nutshell, we perform non-parametric level set estimation, partitioning the data into two sets (ID vs. OOD) based on the deep $k$-nearest neighbor distance.  \method~offers compelling advantages of being: (1) \textbf{distributional assumption free}, (2) \textbf{OOD-agnostic} (\emph{i.e.}, the distance threshold is estimated on the ID data only, and does not rely on information of unknown data),  (3) \textbf{easy-to-use} (\emph{i.e.}, no need to calculate the inverse of the covariance matrix which can be numerically unstable), and (4) \textbf{model-agnostic} (\emph{i.e.}, the testing procedure is applicable to different model architectures and training losses).

Our exploration leads to both empirical effectiveness (Section~\ref{sec:exp} \& \ref{sec:discussion}) and theoretical justification (Section~\ref{sec:theory}). By studying the role of representation space, we show that a compact and normalized feature space is the key to the success of the nearest neighbor approach for OOD detection.  Extensive experiments show that \method~outperforms the parametric approach, and scales well to the large-scale dataset. Computationally, modern implementations of approximate nearest neighbor
search allow us to do this in milliseconds even when the database contains billions of images~\citep{faiss}. On a challenging ImageNet OOD detection benchmark~\cite{huang2021mos}, our KNN-based approach achieves superior performance under a similar inference speed as the baseline methods. The overall simplicity and effectiveness of \method~make it appealing for real-world applications. 
We summarize our contributions below:

\begin{enumerate}

\item We present the first study exploring and demonstrating the efficacy of non-parametric density estimation with nearest neighbors for OOD detection---a simple, flexible yet overlooked approach in literature. We hope our work draws attention to the strong promise of the non-parametric approach, which obviates data assumption on the feature space.

\item We demonstrate the superior performance of the KNN-based method on several OOD detection benchmarks, different model architectures (including CNNs and ViTs), and different training losses. Under the same model trained on ImageNet-1k, our method substantially reduces the false positive rate (FPR@TPR95) by \textbf{24.77}\% compared to a strong baseline {SSD+}~\citep{2021ssd}, which uses a parametric approach (\emph{i.e.}, Mahalanobis distance~\citep{lee2018simple}) for detection.

\item We offer new insights on the key components to make \method~effective in practice, including feature normalization and a compact representation space. Our findings are supported by extensive ablations and experiments. We believe these insights are valuable to the community in carrying out future research.

\item We provide theoretical analysis, showing that KNN-based OOD detection can reject inputs equivalent to the Bayes optimal estimator. By modeling the nearest neighbor distance in
the feature space, our theory (1) directly connects to our method which also operates in the feature space, and (2) complements our experiments by considering the universality of OOD data. 

\end{enumerate}

%%%%%%%%%%%%%%%%%%%%%%%%%%%%%%%%%%%%%%%%%%%%%%%%%%%%%%%%%%%%%%
%%%%%%%%%%%%%%%%%%%%%%%%  PRELIM SECTION %%%%%%%%%%%%%%%%%%%%%
%%%%%%%%%%%%%%%%%%%%%%%%%%%%%%%%%%%%%%%%%%%%%%%%%%%%%%%%%%%%%%

\section{Preliminaries}
\label{sec:prelim}

We consider supervised multi-class classification, where $\mathcal{X}$ denotes the input space and $\mathcal{Y}=\{1,2,...,C\}$ denotes the label space. The training set $\mathbb{D}_{in} = \{(\bx_i, y_i)\}_{i=1}^n$ is drawn \emph{i.i.d.} from the joint data distribution $P_{\mathcal{X}\mathcal{Y}}$. Let $\mathcal{P}_\text{in}$ denote the marginal distribution on $\mathcal{X}$. Let $f: \calX \mapsto \mathbb{R}^{|\calY|}$ be a neural network trained on samples drawn from $P_{\mathcal{X}\mathcal{Y}}$ to output a logit vector, which is used to predict the label of an input sample.

\textbf{Out-of-distribution detection} When deploying a machine model in the real world, {a reliable classifier should not only accurately classify known in-distribution (ID) samples, but also identify as ``unknown'' any OOD input}. This can be achieved by having an OOD detector, in tandem with the classification model $f$. 

OOD detection can be formulated as a binary classification problem. At test time, the goal of OOD detection is to decide whether a sample $\bx \in \mathcal{X}$ is from $\mathcal{P}_\text{in}$ (ID) or not (OOD). The decision can be made via a level set estimation:
\vspace{-0.1cm}
\begin{align*}
\label{eq:threshold}
	G_{\lambda}(\*x)=\begin{cases} 
      \text{ID} & S(\bx)\ge \lambda \\
      \text{OOD} & S(\bx) < \lambda 
   \end{cases},
\end{align*}
where samples with higher scores $S(\bx)$ are classified as ID and vice versa, and  $\lambda$ is the threshold. In practice, OOD is often defined by a distribution that simulates unknowns encountered during deployment time, such as samples from an irrelevant distribution {whose label set has no intersection with $\mathcal{Y}$ and therefore should not be predicted by the model}.

%%%%%%%%%%%%%%%%%%%%%%%%%%%%%%%%%%%%%%%%%%%%%%%%%%%%%%%%%%%%%%
%%%%%%%%%%%%%%%%%%%%%%  METHOD SECTION %%%%%%%%%%%%%%%%%%%%%%%
%%%%%%%%%%%%%%%%%%%%%%%%%%%%%%%%%%%%%%%%%%%%%%%%%%%%%%%%%%%%%%
%
%

\section{Deep Nearest Neighbor for OOD detection}
\label{sec:knn}

In this section, we describe our approach using the deep $k$-Nearest Neighbor (\method) for OOD detection. We illustrate our approach in Figure~\ref{fig:umap}, which at a high level, can be categorized as a distance-based method. Distance-based methods leverage feature embeddings
extracted from a model and operate under the assumption that the test OOD samples are relatively far away from the ID data. Previous distance-based OOD detection methods employed parametric density estimation and modeled the feature embedding space as a mixture of multivariate Gaussian distributions~\citep{lee2018simple}. However, such an approach makes a strong distributional assumption of the learned feature space, which may not necessarily hold\footnote{We verified this by performing the Henze-Zirkler multivariate normality test~\citep{henze1990class} on the embeddings. The testing results show that the feature vectors for each class are not normally distributed at the significance level of 0.05.}.

In this paper, we instead explore the efficacy of \emph{non-parametric density estimation using nearest neighbors} for OOD detection. Despite the simplicity, KNN approach is not systematically explored or compared in most
current OOD detection papers. Specifically, we compute the  $k$-th nearest neighbor distance between the
embedding of each test image and the training set, and use
a simple threshold-based criterion to determine if an input
is OOD or not. Importantly, we use the normalized penultimate feature $\bz= \phi(\bx) / \lVert \phi(\bx) \rVert_2$ for OOD detection, where $\phi: \mathcal{X} \mapsto \mathbb{R}^{m}$ is a feature encoder. Denote the embedding set of training data as $\mathbb{Z}_n = (\bz_1, \bz_2, ..., \bz_n )$. During testing, we derive the normalized feature vector $\bz^*$ for a test sample $\bx^*$, and calculate the Euclidean distances $\lVert\bz_i - \bz^*\rVert_2$ with respect to embedding vectors $\bz_i \in \mathbb{Z}_n$. We reorder $\mathbb{Z}_n$ according to the increasing distance $\lVert\bz_i - \bz^*\rVert_2$. 
Denote the reordered data sequence as $\mathbb{Z}_n' = (\bz_{(1)}, \bz_{(2)}, ..., \bz_{(n)})$. The decision function for OOD detection is given by: 
\begin{equation*}
    G(\bz^*;k) = \mathbf{1}\{-r_k(\bz^*) \ge \lambda\},
\end{equation*} 
where $r_k(\bz^*) = \lVert\bz^* - \bz_{(k)}\rVert_2$ is the distance to the $k$-th nearest neighbor ($k$-NN) and $\mathbf{1}\{\cdot\}$ is the indicator function. The threshold $\lambda$ is typically chosen so that a high fraction of ID data (\emph{e.g.,} 95\%) is correctly classified. The threshold does not depend on OOD data.

\begin{algorithm}[t]
\begin{algorithmic}
   \STATE {\textbf{Input:}} Training dataset $\mathbb{D}_{in}$, pre-trained neural network encoder $\phi$, test sample $\bx^*$, threshold $\lambda$\\
   
   \STATE For $\bx_i$ in the training data $\mathbb{D}_{in}$,  collect feature vectors $\mathbb{Z}_n = (\bz_1, \bz_2, ..., \bz_n )$

   \STATE \textbf{Testing Stage}: 
        \STATE Given a test sample, we calculate feature vector $\bz^* = \phi(\bx^*) / \lVert \phi(\bx^*) \rVert_2$
        \STATE Reorder $\mathbb{Z}_n$ according to the increasing value of $\lVert\bz_i - \bz^*\rVert_2$ as $\mathbb{Z}_n' = (\bz_{(1)}, \bz_{(2)}, ..., \bz_{(n)})$
    \STATE \textbf{Output: } OOD detection decision $\mathbf{1}\{-\lVert\bz^* - \bz_{(k)}\rVert_2 \ge \lambda\}$
\end{algorithmic}
\caption{OOD Detection with Deep Nearest Neighbors}
\label{alg}
\end{algorithm}

%%%%%%%%%%%%%%%%%%%%%%%  TABLE CIFAR %%%%%%%%%%%%%%%%%%%%%%%

\begin{table*}[t]
\caption{ \textbf{Results on CIFAR-10.} Comparison with competitive OOD detection methods. All methods are based on a discriminative model trained on {ID data only}, without using outlier data.  $\uparrow$ indicates larger values are better and vice versa. } 
\centering
\scalebox{0.75}{

\begin{tabular}{llllllllllllll} \toprule
\multicolumn{1}{c}{\multirow{3}{*}{\textbf{Method}}} & \multicolumn{10}{c}{\textbf{OOD Dataset}} & \multicolumn{2}{c}{\multirow{2}{*}{\textbf{Average}}} & \multirow{3}{*}{\textbf{ID ACC}} \\
\multicolumn{1}{c}{} & \multicolumn{2}{c}{\textbf{SVHN}} & \multicolumn{2}{c}{\textbf{LSUN }} &  \multicolumn{2}{c}{\textbf{iSUN}} & \multicolumn{2}{c}{\textbf{Texture}} & \multicolumn{2}{c}{\textbf{Places365}} & \multicolumn{2}{c}{} & \\
\multicolumn{1}{c}{} & \textbf{FPR$\downarrow$} & \textbf{AUROC$\uparrow$} & \textbf{FPR$\downarrow$} & \textbf{AUROC$\uparrow$} &  \textbf{FPR$\downarrow$} & \textbf{AUROC$\uparrow$} & \textbf{FPR$\downarrow$} & \textbf{AUROC$\uparrow$} & \textbf{FPR$\downarrow$} & \textbf{AUROC$\uparrow$} & \textbf{FPR$\downarrow$} & \textbf{AUROC$\uparrow$} & \\ \midrule  &\multicolumn{13}{c}{\textbf{Without Contrastive Learning}}          \\
MSP & 59.66 & 91.25 & 45.21 & 93.80 & 54.57 & 92.12 & 66.45 & 88.50 & 62.46 & 88.64 & 57.67 & 90.86 & 94.21 \\
ODIN & 53.78 & 91.30 & 10.93  & 97.93 & 28.44 & 95.51 & 55.59 & 89.47 & 43.40 & 90.98 & 38.43 & 93.04 & 94.21 \\
Energy & 54.41 & 91.22 & 10.19 & 98.05 & 27.52 & 95.59 & 55.23 & 89.37 & 42.77 & 91.02 & 38.02 & 93.05 & 94.21 \\
GODIN & 18.72 & 96.10 & 11.52  & 97.12 & 30.02 & 94.02 & 33.58 & 92.20 & 55.25 & 85.50 & 29.82 & 92.97 & 93.64 \\ 
Mahalanobis & 9.24  & 97.80 & 67.73 & 73.61 & 6.02  & 98.63 & 23.21 & 92.91 & 83.50 & 69.56 & 37.94 & 86.50 & 94.21 \\ 
% KNN (ours) & 24.53 & 95.96 & 25.29 & 95.69 & 25.55 & 95.26 & 27.57 & 94.71 & 50.90 & 89.14 & 30.77 & 94.15 & 94.21\\ \midrule
KNN (ours) & 27.97 & 95.48 & 18.50 & 96.84 & 24.68 & 95.52  & 26.74 & 94.96 & 47.84 & 89.93  & 29.15 & 94.55 & 94.21\\ \midrule
&\multicolumn{13}{c}{\textbf{With Contrastive Learning}}          \\
CSI & 37.38 & 94.69 & 5.88  & 98.86 & 10.36 & 98.01 & 28.85 & 94.87 & 38.31 & 93.04 & 24.16 & 95.89 &  94.38 \\
SSD+ & 1.51  & 99.68 & 6.09  & 98.48 & 33.60 & 95.16 & 12.98 & 97.70 & 28.41 & 94.72 & 16.52 & 97.15 & \textbf{95.07} \\
KNN+ (ours) & 2.42  & 99.52 & 1.78  & 99.48 & 20.06 & 96.74 & 8.09  & 98.56 & 23.02 & 95.36 & \textbf{11.07} & \textbf{97.93} & \textbf{95.07} \\ \bottomrule
\end{tabular}
}
\label{tab:cifar_main}
\end{table*}

%%%%%%%%%%%%%%%%%%%%%%%  TABLE CIFAR %%%%%%%%%%%%%%%%%%%%%%%

We summarize our approach in Algorithm~\ref{alg}. Noticeably, KNN-based OOD detection offers several compelling advantages:
\begin{enumerate}
\vspace{-0.2cm}
     \item \textbf{Distributional assumption free}: Non-parametric nearest neighbor approach does not impose distributional assumptions about the underlying feature space. KNN therefore provides stronger flexibility and generality, and is applicable even when the feature space does not conform to the mixture of Gaussians. 
       \item \textbf{OOD-agnostic}: The testing procedure does not rely on the information of unknown data. The distance threshold is estimated on the ID data only. 
     \item \textbf{Easy-to-use}: Modern implementations of approximate nearest neighbor search
allow us to do this in milliseconds even when the database contains billions of images~\citep{faiss}. In contrast, Mahalanobis distance requires calculating the inverse of the covariance matrix, which can be numerically unstable. 
\vspace{-0.2cm}
   \item \textbf{Model-agnostic}: The testing procedure applies to a variety of model architectures, including CNNs and more recent Transformer-based  ViT models~\citep{dosovitskiy2020image}. Moreover, we will show that \method~is agnostic to the training procedure as well, and is compatible with models trained under different loss functions (\emph{e.g.}, cross-entropy loss and contrastive loss). 
\end{enumerate}
  We proceed to show the efficacy of the KNN-based OOD detection approach in Section~\ref{sec:exp}.

%%%%%%%%%%%%%%%%%%%%%%%%%%%%%%%%%%%%%%%%%%%%%%%%%%%%%%%%%%%%%%
%%%%%%%%%%%%%%%%%%%%  EXPERIMENT SECTION %%%%%%%%%%%%%%%%%%%%%
%%%%%%%%%%%%%%%%%%%%%%%%%%%%%%%%%%%%%%%%%%%%%%%%%%%%%%%%%%%%%%

\section{Experiments}
\label{sec:exp}

The goal of our experimental evaluation is to answer the
following questions: (1) How does \method~fare against the parametric counterpart such as Mahalanobis distance for OOD detection? (2) Can \method~scale to a more challenging task when the training data is large-scale (\emph{e.g.}, ImageNet)? (3) Is KNN-based OOD detection effective under different model architectures and training objectives? (4) How do various design choices affect the performance?

\vspace{-0.3cm}
\paragraph{Evaluation metrics} 
We report the following metrics: (1) the false positive rate (\text{FPR}95) of OOD samples when the true positive rate of ID samples is at 95\%, (2) the area under the receiver operating characteristic curve (AUROC), (3) ID classification accuracy (ID ACC), and (4) per-image inference time (in milliseconds, averaged across test images).

\vspace{-0.3cm}
\paragraph{Training losses} In our experiments, we aim to show that KNN-based OOD detection is agnostic to the training procedure, and is compatible with models trained under different losses. We consider two types of loss functions, with and without contrastive learning respectively. We employ (1) cross-entropy loss which is the most commonly used training objective in classification, and (2) supervised contrastive learning (SupCon)~\citep{2020supcon}--- the latest development for representation learning, which leverages the label information by aligning samples belonging to the same class in the embedding space. 

\vspace{-0.3cm}
\paragraph{Remark on the implementation} 
All of the experiments are
based on PyTorch~\citep{pytorch}. Code is made publicly available online. We use Faiss~\citep{faiss}, a library for efficient nearest neighbor search. Specifically, we use \texttt{faiss.IndexFlatL2} as the indexing method with Euclidean distance. In practice, we pre-compute the embeddings for all images and store
them in a key-value map to make KNN search efficient. The embedding vectors for ID data only need to be extracted once after the training is completed. 

\subsection{Evaluation on Common Benchmarks}
\label{sec:common_benchmark}

\paragraph{Datasets}
We begin with the CIFAR benchmarks that are routinely used in literature. We use the standard split with 50,000 training images and 10,000 test images. We evaluate the methods on common OOD datasets: \texttt{Textures}~\citep{cimpoi2014describing}, \texttt{SVHN}~\citep{netzer2011reading}, \texttt{Places365}~\citep{zhou2017places}, \texttt{LSUN-C}~\citep{yu2015lsun}, and \texttt{iSUN}~\citep{xu2015turkergaze}. All images are of size $32\times 32$.

\vspace{-0.3cm}
\paragraph{Experiment details}
We use ResNet-18 as the backbone for CIFAR-10. Following the original settings in \citeauthor{2020supcon}, models with \texttt{SupCon} loss are trained for 500 epochs, with the batch size of $1024$. The temperature $\tau$ is $0.1$. The dimension of the penultimate feature where we perform the nearest neighbor search is 512. The dimension of the projection head is 128. We use the cosine annealing learning rate~\citep{loshchilov2016sgdr} starting at 0.5. We use $k=50$ for CIFAR-10 and $k=200$ for CIFAR-100, which is selected from $k=\{1,10,20,50,100,200,500,1000,3000,5000\}$ using the validation method in ~\citep{hendrycks2018deep}. 
We train the models using stochastic gradient descent with momentum 0.9, and weight decay  $10^{-4}$. The model without contrastive learning is trained for 100 epochs. The start learning rate is 0.1 and decays by a factor of 10 at epochs 50, 75, and 90 respectively.

\begin{table*}[t]
\centering
\caption{Evaluation (FPR95) on hard OOD detection tasks. Model is trained on CIFAR-10 with SupCon loss. }
\scalebox{0.9}{
\begin{tabular}{ccccc} \toprule
    & \textbf{LSUN-FIX} & \textbf{ImageNet-FIX} & \textbf{ImageNet-R} & \textbf{C-100} \\ \midrule
{SSD+} & 29.86      & 32.26          &  45.62             & 45.50     \\
{\methodplus~(Ours)} & \textbf{21.52}       & \textbf{25.92}          & \textbf{29.92}            & \textbf{38.83} \\ \bottomrule 
\end{tabular}}
\label{tab:hardood}
\end{table*}

\paragraph{Nearest neighbor distance achieves superior performance}
We present results in Table~\ref{tab:cifar_main}, where non-parametric KNN approach shows favorable performance. Our comparison covers an extensive collection of competitive methods in the literature. For clarity, we divide the baseline methods into two categories: trained with and without contrastive losses. Several baselines derive OOD scores from a model trained with common softmax cross-entropy (CE) loss, including \texttt{MSP}~\citep{Kevin}, \texttt{ODIN}~\citep{liang2018enhancing}, \texttt{Mahalanobis}~\citep{lee2018simple}, and \texttt{Energy}~\citep{liu2020energy}. \texttt{GODIN}~\citep{hsu2020generalized} is trained using a DeConf-C loss, which does not involve contrastive loss either. 
For methods involving contrastive losses, we use the same network backbone architecture and embedding dimension, while only varying the training objective. These methods include 
\texttt{CSI}~\citep{tack2020csi} and
\texttt{SSD+}~\citep{2021ssd}. For terminology clarity, \method~refers to our method trained with CE loss, and \methodplus~refers to the variant trained with SupCon loss. We highlight two groups of comparisons: 
\begin{itemize}
\vspace{-0.2cm}
\item \textbf{\method~vs. Mahalanobis} (without contrastive learning): Under the \emph{same} model trained with cross-entropy (CE) loss, our method achieves an average FPR95 of 29.15\%, compared to that of Mahalanobis distance 37.94\%. The performance gain precisely demonstrates the advantage of \method~over the parametric method Mahalanobis distance.
\vspace{-0.2cm}
\item \textbf{\methodplus~vs. SSD+} (with contrastive loss):   \methodplus~and \texttt{SSD+}  are fundamentally different in OOD detection mechanisms, despite both benefit from the contrastively learned representations.  \texttt{SSD+} modeled the feature embedding space as a multivariate Gaussian distribution for each class, and use Mahalanobis distance~\cite{lee2018simple} for OOD detection. Under the \emph{same} model trained with Supervised Contrastive Learning (SupCon) loss, our method with the nearest neighbor distance reduces the average FPR95 by ${5.45}\%$, which is a relatively \textbf{32.99}\% reduction in error. It further suggests the advantage of using nearest neighbors without making any distributional assumptions on the feature embedding space.  

\end{itemize}
	\vspace{-0.2cm}
The above comparison suggests that the nearest neighbor approach is compatible with models trained both with and without contrastive learning. In addition, \method~is also simpler to use and implement than \texttt{CSI}, which relies on sophisticated data augmentations and ensembling in testing. Lastly, as a result of the improved embedding quality, the ID accuracy of the model trained with \texttt{SupCon} loss is improved by ${0.86}\%$ on CIFAR-10 and 2.45\% on ImageNet compared to training with the \texttt{CE} loss. 
Due to space constraints, we provide results on  DenseNet~\citep{huang2017densely} in Appendix~\ref{sec:other_arc}.

\paragraph{Contrastively learned representation helps} While contrastive learning has been extensively studied in recent literature, {its role remains untapped when coupled with a non-parametric approach} (such as nearest neighbors) for OOD detection. We examine the effect of using supervised contrastive loss for KNN-based OOD detection. We provide both qualitative and quantitative evidence, highlighting advantages over the standard softmax cross-entropy (\texttt{CE}) loss. (1) We visualize the learned feature embeddings in Figure~\ref{fig:umap} using UMAP~\citep{mcinnes2018umap-software}, where the colors encode different class labels. A salient observation is that the representation with \texttt{SupCon} is more distinguishable and compact than the representation obtained from the \texttt{CE} loss. The high-quality embedding space indeed confers benefits for KNN-based OOD detection. (2) Beyond visualization, we also quantitatively compare the performance of KNN-based OOD detection using embeddings trained with \texttt{SupCon} vs \texttt{CE}. As shown in 
Table~\ref{tab:cifar_main}, \methodplus~with contrastively learned representations reduces the FPR95 on all test OOD datasets compared to using embeddings from the model trained with \texttt{CE} loss.

\vspace{-0.2cm}
\paragraph{Comparison with other non-parametric methods } In Table~\ref{tab:nonparam}, we compare the nearest neighbor approach with other non-parametric methods. For a fair comparison, we use the same embeddings trained with \texttt{SupCon} loss. Our comparison  covers  an  extensive  collection of outlier detection methods in literature including: \texttt{IForest}~\citep{liu2008iforest}, \texttt{OCSVM}~\citep{bernhard2001ocsvm}, 
\texttt{LODA}~\citep{2016loda}, 
\texttt{PCA}~\citep{shyu2003pca}, and
\texttt{LOF}~\citep{breunig2000lof}. The parameter setting for these methods is available in Appendix~\ref{sec:config}. We show that \methodplus~outperforms alternative non-parametric methods by a large margin.

%%%%%%%%%%%%%%%%%%%%%%%  TABLE non-param %%%%%%%%%%%%%%%%%%%%%%%
\begin{table}[h]
\centering
\caption{Comparison with other non-parametric methods. Results are averaged across all test OOD datasets. Model is trained on CIFAR-10.}
\vspace{0.2cm}
\scalebox{0.95}{
\begin{tabular}{lll} \toprule
 & \textbf{FPR95}$\downarrow$ & \textbf{AUROC}$\uparrow$ \\ \midrule
IForest~\citep{liu2008iforest} & 65.49 & 76.98 \\
OCSVM~\citep{bernhard2001ocsvm}   & 52.27 & 65.16 \\
LODA~\citep{2016loda}    & 76.38 & 62.59 \\
PCA~\citep{shyu2003pca}     & 37.26 & 83.13 \\
LOF~\citep{breunig2000lof}     & 40.06 & 93.47 \\
\methodplus~(ours)    & \textbf{11.07} & \textbf{97.93} \\ \bottomrule
\end{tabular}}
\label{tab:nonparam}
\end{table}

%%%%%%%%%%%%%%%%%%%%%%%  TABLE non-param %%%%%%%%%%%%%%%%%%%%%%%

%%%%%%%%%%%%%%%%%%%%%%%  TABLE ImageNet %%%%%%%%%%%%%%%%%%%%%%%
\begin{table*}[t]
\centering
\caption{\textbf{Results on ImageNet}. All methods are based on a model trained on ID data only (ImageNet-1k~\citep{deng2009imagenet}). We report the OOD detection performance, along with the per-image inference time.}
\scalebox{0.75}{ 
\begin{tabular}{lllllllllllll} \toprule
\multirow{4}{*}{\textbf{Methods}} & \multirow{4}{*}{\begin{tabular}[c]{@{}l@{}}\textbf{Inference} \\ \textbf{time (ms)}\end{tabular}} & \multicolumn{8}{c}{\textbf{OOD Datasets}} & \multicolumn{2}{c}{\multirow{2}{*}{\textbf{Average}}} & \multirow{4}{*}{\textbf{ID ACC}}\\
 & & \multicolumn{2}{c}{\textbf{iNaturalist}} & \multicolumn{2}{c}{\textbf{SUN}} & \multicolumn{2}{c}{\textbf{Places}} & \multicolumn{2}{c}{\textbf{Textures}} & \multicolumn{2}{c}{} \\
 & & \textbf{FPR95} & \textbf{AUROC} & \textbf{FPR95} & \textbf{AUROC} & \textbf{FPR95} & \textbf{AUROC} & \textbf{FPR95} & \textbf{AUROC} & \textbf{FPR95} & \textbf{AUROC} \\ 
 & & $\downarrow$ & $\uparrow$ & $\downarrow$ & $\uparrow$ & $\downarrow$ & $\uparrow$ & $\downarrow$ & $\uparrow$ & $\downarrow$ & $\uparrow$ \\ \midrule
 &\multicolumn{10}{c}{\textbf{Without Contrastive Learning}}          \\
MSP & 7.04 & 54.99 & 87.74 & 70.83 & 80.86 & 73.99 & 79.76 & 68.00 & 79.61 & 66.95 & 81.99 & 75.08 \\
ODIN & 7.05 & 47.66 & 89.66 & 60.15 & 84.59 & 67.89 & 81.78 & 50.23 & 85.62 & 56.48 & 85.41 & 75.08 \\
Energy & 7.04 & 55.72 & 89.95 & 59.26 & 85.89 & 64.92 & 82.86 & 53.72 & 85.99 & 58.41 & 86.17 & 75.08 \\
GODIN & 7.04 & 61.91 & 85.40 & 60.83 & 85.60 & 63.70 & 83.81 & 77.85 & 73.27 & 66.07 & 82.02 & 70.43 \\
Mahalanobis &35.83 & 97.00 & 52.65 & 98.50 & 42.41 & 98.40 & 41.79 & 55.80 & 85.01 & 87.43 & 55.47 & 75.08 \\
KNN ($\alpha=100\%$) & 10.31 & 59.77 & 85.89 & 68.88 & 80.08 & 78.15 & 74.10 & 10.90 & 97.42 & 54.68 & 84.37 & 75.08 \\
KNN ($\alpha=1\%$) & 7.04 & 59.08 & 86.20 & 69.53 & 80.10 & 77.09 & 74.87 & 11.56 & 97.18 & 54.32 & 84.59 & 75.08\\
\hline
&\multicolumn{10}{c}{\textbf{With Contrastive Learning}}          \\
SSD+ & 28.31 & 57.16 & 87.77 & 78.23 & 73.10 & 81.19 & 70.97 & 36.37 & 88.52 & 63.24 & 80.09 & \textbf{79.10}\\
KNN+ ($\alpha=100\%$) & 10.47 & 30.18 & 94.89 & 48.99 & 88.63 & 59.15 & 84.71 & 15.55 & 95.40 & \textbf{38.47} & \textbf{90.91} & \textbf{79.10} \\
KNN+ ($\alpha=1\%$) & 7.04 & 30.83 & 94.72 & 48.91 & 88.40 & 60.02 & 84.62 & 16.97 & 94.45 & 39.18 & 90.55 & \textbf{79.10} \\ \bottomrule
\end{tabular}}
\label{tab:imagenet_main}
\end{table*}
%%%%%%%%%%%%%%%%%%%%%%%  TABLE ImageNet %%%%%%%%%%%%%%%%%%%%%%%

%%%%%%%%%%%%%%%%%%%%%%% FIGURE K-FPR %%%%%%%%%%%%%%%%%%%%
\begin{figure*}[tb]
	\begin{center}
		\includegraphics[width=1.0\linewidth]{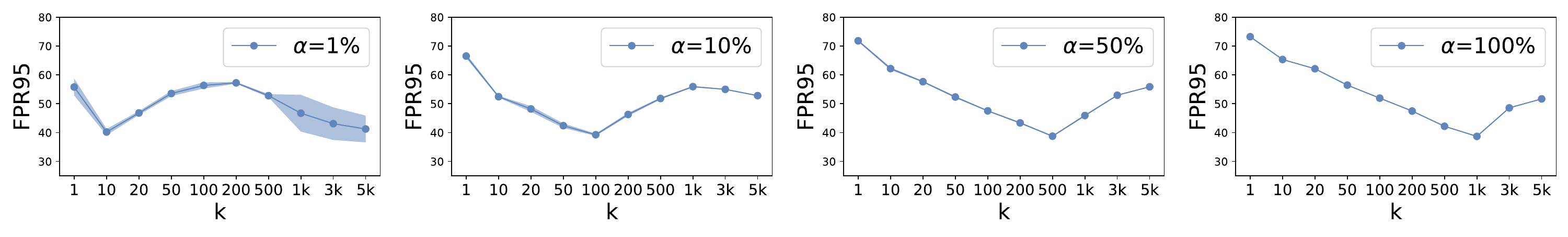}
	\end{center}
	\vspace{-0.4cm}
	\caption{Comparison with the effect of different $k$ and sampling ratio $\alpha$. We report an average FPR95 score over four test OOD datasets. The variances are estimated across 5 different random seeds. The solid blue line represents the averaged value across all runs and the shaded blue area represents the standard deviation. Note that the full ImageNet dataset ($\alpha=100\%$) has 1000 images per class. }
	\label{fig:k_fpr}
\end{figure*}
%%%%%%%%%%%%%%%%%%%%%%% FIGURE K-FPR %%%%%%%%%%%%%%%%%%%%

\paragraph{Evaluations on hard OOD tasks} Hard OOD samples are particularly
challenging to detect. To test the limit of the  non-parametric KNN approach, we follow CSI~\citep{tack2020csi} and evaluate on several hard OOD datasets: LSUN-FIX, ImageNet-FIX, ImageNet-R, and CIFAR-100. The results are summarized in Table~\ref{tab:hardood}. Under the same model, {\methodplus~consistently outperforms \texttt{SSD+}}.

%%%%%%%%%%%%%%%%%%%%%%%  Figure  Ablation Many %%%%%%%%%%%%%%%%%%%%%%%

\begin{figure*}[t]
	\begin{center}
		\includegraphics[width=1.\linewidth]{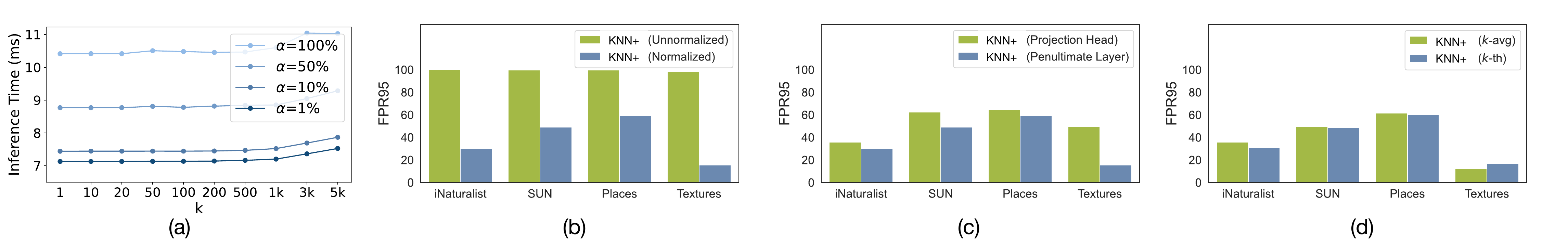}
	\end{center}
	\caption{\small  \textbf{Ablation results.} In (a), we compare the inference speed (per-image) using different $k$ and sampling ration $\alpha$. For (b) (c) (d), the FPR95 value is reported over all test OOD datasets. Specifically, (b) compares the effect of using normalization in the penultimate layer feature vs. without normalization, (c) compares using features in the penultimate layer feature vs the projection head, and (d) compares the OOD detection performance using  $k$-th and averaged $k$ ($k$-avg) nearest neighbor distance. }
	\label{fig:abl_many}
\end{figure*}

%%%%%%%%%%%%%%%%%%%%%%%  Figure  Ablation Many %%%%%%%%%%%%%%%%%%%%%%%

\subsection{Evaluation on Large-scale ImageNet Task}
\label{sec:imagenet}

We evaluate on a large-scale OOD detection task based on ImageNet~\citep{deng2009imagenet}. Compared to the CIFAR benchmarks above, the ImageNet task is more challenging due to a large amount of training data. Our goal is to verify \method's performance benefits and whether it scales computationally with millions of samples.

\paragraph{Setup} We use a ResNet-50 backbone~\citep{he2016identity} and train on ImageNet-1k~\citep{deng2009imagenet} with resolution $224 \times 224$. Following the experiments in \citeauthor{2020supcon}, models with \texttt{SupCon} loss are trained for 700 epochs, with a batch size of $1024$. The temperature $\tau$ is $0.1$. The dimension of the penultimate feature where we perform the nearest neighbor search is 2048. The dimension of the project head is 128. We use the cosine learning rate~\citep{loshchilov2016sgdr} starting at 0.5. We train the models using stochastic gradient descent with momentum 0.9, and weight decay  $10^{-4}$. We use $k=1000$ which follows the same validation procedure as before. When randomly sampling $\alpha\%$ training data for nearest neighbor search, $k$  is scaled accordingly to $1000 \cdot \alpha\%$.

Following the ImageNet-based OOD detection benchmark in MOS~\citep{huang2021mos}, we evaluate on four test OOD datasets that are subsets of: \texttt{Places365}~\citep{zhou2017places}, \texttt{Textures}~\citep{cimpoi2014describing}, \texttt{iNaturalist}~\citep{inat}, and \texttt{SUN}~\citep{sun} with non-overlapping categories \emph{w.r.t.} ImageNet. The evaluations span a diverse range of domains including fine-grained images, scene images, and textural images.

\vspace{-0.2cm}
\paragraph{Nearest neighbor approach achieves superior performance without compromising the inference speed} In Table~\ref{tab:imagenet_main}, we compare our approach with OOD detection methods that are competitive in the literature. The baselines are the same as what we described in Section~\ref{sec:common_benchmark} except for \texttt{CSI}\footnote{The training procedure of \texttt{CSI} is computationally  prohibitive on ImageNet, which takes three months on 8 Nvidia 2080Tis.}. We report both  OOD detection performance and the inference time (measured by milliseconds). We highlight three trends: (1) \methodplus~outperforms the best baseline by \textbf{18.01}\% in FPR95. (2) Compared to \texttt{SSD+}, \methodplus~substantially reduces the FPR95 by $\textbf{24.77}\%$ averaged across all test sets. The limiting performance of \texttt{SSD+} is due to the increased size of label space and data complexity, which makes the class-conditional Gaussian assumption less viable. In contrast, our non-parametric method does not suffer from this issue, and can better estimate the density of the complex distribution for OOD detection. (3) \methodplus~achieves strong performance with a comparable inference speed as the baselines. In particular, we show that performing nearest neighbor distance estimation with only $1\%$ randomly sampled training data can yield a similar performance as using the full dataset.

\paragraph{Nearest neighbor approach is competitive on ViT} Going beyond convolutional neural networks, we show in Table~\ref{tab:vit} that the nearest neighbor approach is effective for transformer-based  ViT model~\citep{dosovitskiy2020image}. We adopt the ViT-B/16 architecture fine-tuned on the ImageNet-1k dataset using cross-entropy loss. Under the same ViT model, our non-parametric KNN method consistently outperforms Mahalanobis. 

\begin{table*}[h]
\centering
\caption{Performance comparison (FPR95) on ViT-B/16 model fine-tuned on ImageNet-1k.}
\scalebox{1.0}{
\begin{tabular}{ccccc} \toprule
    & \textbf{iNaturalist} & \textbf{SUN} & \textbf{Places} & \textbf{Textures} \\ \midrule
{Mahalanobis (parametric)} & 17.56 & 80.51 & 84.12 & 70.51  \\
{KNN (non-parametric)} & \textbf{7.30} & \textbf{48.40} & \textbf{56.46} & \textbf{39.91} \\ \bottomrule 
\end{tabular}}
\label{tab:vit}
\end{table*}

\section{A Closer Look at KNN-based OOD Detection}
We provide further analysis and ablations to understand the behavior of KNN-based OOD detection. All the ablations are based on the ImageNet model trained with SupCon loss (same as in Section~\ref{sec:imagenet}).

\label{sec:discussion}

\paragraph{Effect of $k$ and sampling ratio}
In Figure~\ref{fig:k_fpr} and Figure~\ref{fig:abl_many} (a), we systematically analyze the effect of $k$ and the dataset sampling ratios $\alpha$. We vary the number of neighbors $k=\{1,10,20,50,100,200,500,1000,3000,5000\}$ and random sampling ratio $\alpha = \{1\%,10\%,50\%,100\%\}$. We note several interesting observations: (1) The optimal OOD detection (measured by FPR95) remains \emph{similar} under different random sampling ratios $\alpha$. (2) The optimal $k$ is consistent with the one chosen by our validation strategy. For example, the optimal $k$ is 1,000 when $\alpha=100\%$; and the optimal $k$ becomes 10 when $\alpha=1\%$. (3) Varying $k$ does not significantly affect the inference speed when $k$ is relatively small (\emph{e.g.}, $k<1000$) as shown in Figure~\ref{fig:abl_many} (a).

\begin{figure}[t]
	\begin{center}
		\includegraphics[width=0.95\linewidth]{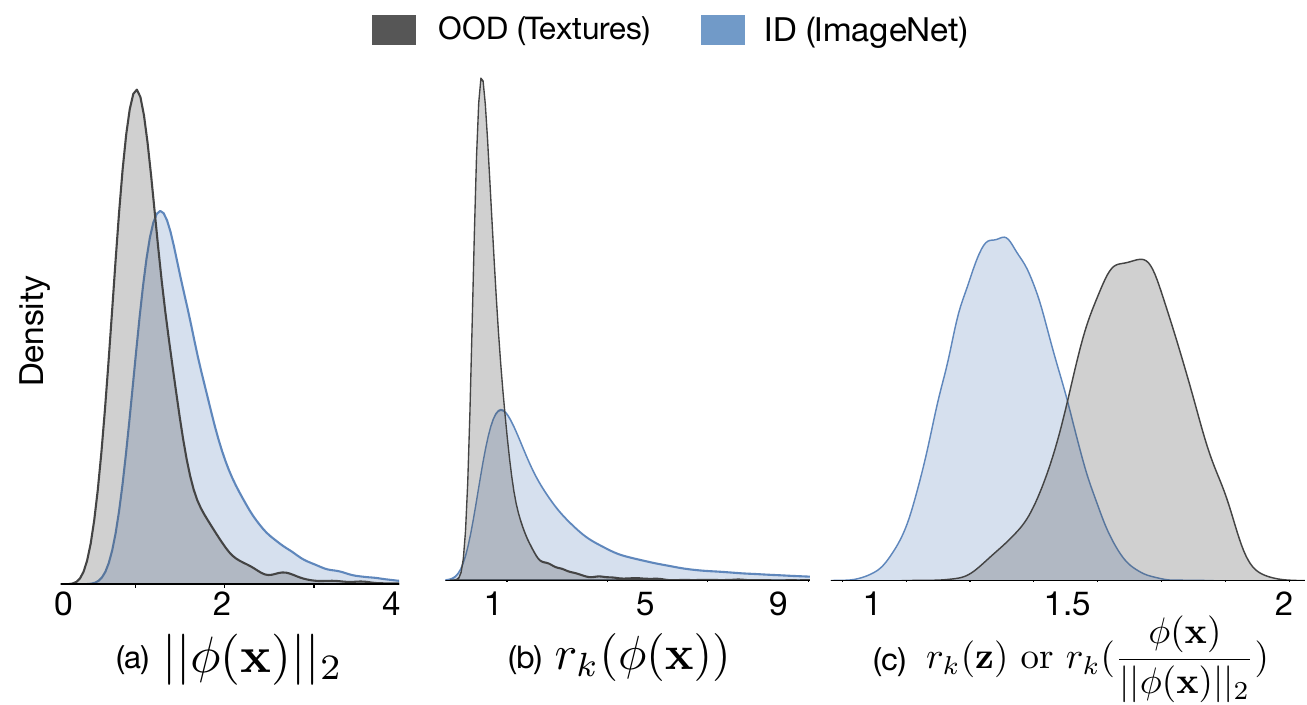}
	\end{center}	
	\vspace{-0.3cm}
	\caption{\small Distribution of (a) the $L_2$-norm of feature embeddings, (b) the $k$-NN distance with the \emph{unnormalized} feature embeddings, and (c) the $k$-NN distance with the \emph{normalized} features. }
	\label{fig:knn_norm}
\vspace{-0.3cm}
\end{figure}

\paragraph{Feature normalization is critical}
In this ablation, we contrast the performance of KNN-based OOD detection with and without feature normalization. The $k$-th NN distance can be derived by  $r_k(\frac{\phi(\bx)}{\lVert(\phi(\bx)\rVert})$ and $r_k(\phi(\bx))$, respectively. 
As shown in Figure~\ref{fig:abl_many} (b), using feature normalization improved the FPR95 drastically by \textbf{61.05}\%, compared to the counterpart without normalization. To better understand this, we look into  the Euclidean distance $r=\lVert u - v \rVert_2$ between two vectors $u$ and $v$. The norm of the feature vector $u$ and $v$ could notably affect the value of the Euclidean distance. Interestingly, recent studies share the observation in Figure~\ref{fig:knn_norm} (a) that the ID data has a larger $L_2$ feature norm than OOD data~\citep{tack2020csi, huang2021importance}. Therefore, the Euclidean distance between ID features can be large (Figure~\ref{fig:knn_norm} (b)).  This contradicts the hope that ID data has a smaller $k$-NN distance than OOD data. 
Indeed, the normalization effectively mitigated this problem, as evidenced in Figure~\ref{fig:knn_norm} (c). Empirically, the normalization plays a key role in the nearest neighbor approach to be successful in OOD detection as shown in Figure~\ref{fig:abl_many} (b).

%%%%%%%%%%%%%%%%%%%%%%%  Figure  Ablation Many %%%%%%%%%%%%%%%%%%%%%%%

%%%%%%%%%%%%%%%%%%%%%%%  Figure  Ablation Many %%%%%%%%%%%%%%%%%%%%%%%

\vspace{-0.1cm}
\paragraph{Using the penultimate layer's feature is better than using the projection head}
In this paper, we follow the convention in \texttt{SSD+}, which uses features from the penultimate layer instead of the projection head. We also verify in Figure~\ref{fig:abl_many} (c) that using the penultimate layer's feature is better than using the projection head on all test OOD datasets. This is likely due to the penultimate layer preserves more information than the projection head, which has much smaller dimensions.

\vspace{-0.1cm}
\paragraph{KNN can be further boosted by activation rectification} We show that \methodplus~can be made stronger with a recent method of activation rectification~\citep{sun2021react}. It was shown that the OOD data can have overly high activations on some feature dimensions, and this rectification is effective in suppressing the values. Empirically, we compare the results in Table~\ref{tab:heu} by using the activation rectification and achieve improved OOD detection performance.

\begin{table}[h]
    \centering
    \caption{Comparison of KNN-based method with and without activation truncation. The ID data is ImageNet-1k. The value is averaged over all test OOD datasets.}
    \label{tab:heu}
    \scalebox{0.9}{
    \begin{tabular}{c|cc}
    \toprule
    Method & {FPR95}\textbf{$\downarrow$} & {AUROC} \textbf{$\uparrow$} \\  \midrule
    KNN+ & 38.47 & 90.91 \\
     KNN+ (w. ReAct~\cite{sun2021react}) & \textbf{26.45} & \textbf{93.76} \\
     \bottomrule
    \end{tabular}}
\end{table}

\vspace{-0.1cm}
\paragraph{Using $k$-th and averaged $k$ nearest nerighbors' distance has similar performance}
We compare two variants for OOD detection: $k$-th nearest neighbor distance vs. averaged $k$ ($k$-avg) nearest neighbor distance. The comparison is shown in Figure~\ref{fig:abl_many} (d), where the average performance (on four datasets) is on par. The reported results are based on the full ID dataset ($\alpha=100\%$) with the optimal $k$ chosen for $k$-th NN and $k$-avg NN respectively. Despite the similar performance, using $k$-th NN distance has a stronger theoretical interpretation, as we show in the next section.

\section{Theoretical Justification}
\label{sec:theory}

In this section, we provide a theoretical analysis of using KNN for OOD detection. By modeling the KNN in the feature space, our theory (1) directly connects to our method which also operates in the feature space, and (2) complements our experiments by considering the universality of OOD data. Our goal here is to analyze the average performance of our algorithm while being OOD-agnostic and training-agnostic.

\paragraph{Setup}
We consider OOD detection task as a special binary classification task, where the negative samples (OOD) are only available in the testing stage. We assume the input is from feature embeddings space $\mathcal{Z}$ and the labeling set $\mathcal{G} = \{0 (\text{OOD}), 1 (\text{ID})\}$. In the inference stage, the testing set $\{(\bz_i, g_i)\}$ is drawn \textit{i.i.d.} from $P_{\mathcal{Z}\mathcal{G}}$. 

Denote the marginal distribution on $\mathcal{Z}$ as $\calP$. We adopt the Huber contamination model~\citep{huber1964} to model the fact that we may encounter both ID and OOD data in test time:
$$
\calP = \varepsilon \calP_{out} + (1 - \varepsilon )\calP_{in},
$$
where $\calP_{in}$ and $\calP_{out}$ are the underlying distributions of feature embeddings for ID and OOD data, respectively, and $\varepsilon$ is a constant controlling the fraction of OOD samples in testing. We use lower case $p_{in}(\bz_i)$ and $p_{out}(\bz_i)$ to denote the probability density function, where $p_{in}(\bz_i) = p(\bz_i| g_i=1)$ and $p_{out}(\bz_i) = p(\bz_i| g_i=0)$. 

A key challenge in OOD detection (and theoretical analysis) is the lack of knowledge on OOD distribution, which can arise universally outside ID data. We thus try to keep our analysis general and reflect the fact that we do not have any strong prior information about OOD. For this reason, we model OOD data with an equal chance to appear outside of the high-density region of ID data, $p_{out}(\bz) =c_0\mathbf{1}\{p_{in}(\bz) < c_1\}$\footnote{In experiments, as it is difficult to simulate the universal OOD, we approximate it by using a diverse yet finite collection of datasets. Our theory is thus complementary to our experiments and captures the universality of OOD data.}. The Bayesian classifier is known as the optimal binary classifier defined by $h_{Bay}(\bz_i) = \mathbf{1}\{p(g_i = 1|\bz_i) \ge \beta\}$\footnote{Note that $\beta$ does not have to be $\frac{1}{2}$ for the Bayesian classifier to be optimal. $\beta$ can be any value 
larger than $\frac{(1-\epsilon)c_1}{(1-\epsilon)c_1 + \epsilon c_0}$ when $\epsilon c_0 \ge (1-\epsilon)c_1$.}, assuming the underlying density function is given. 

Without such oracle information, our method applies $k$-NN as the distance measure which acts as a probability density estimation, and thus provides the decision boundary based on it. Specifically, \method's hypothesis class $\mathcal{H}$ is given by $\{h:h_{\lambda, k, \mathbb{Z}_n}(\bz_i) = \mathbf{1}\{-r_k(\bz_i) \ge \lambda\}\}$, where $r_k(\bz_i)$ is the distance to the $k$-th nearest neighbor (\emph{c.f.} Section~\ref{sec:knn}).

\paragraph{Main result} We show that our KNN-based OOD detector can reject inputs equivalent to the estimated Bayesian binary decision function. A small KNN distance $r_k(\bz_i)$ directly translates into a high probability of being ID, and vice versa. We depict this in the following Theorem.

\begin{theorem} 
\label{th:knn2bayes}
With the setup specified above, if $\hat{p}_{out}(\bz_i) = \hat{c}_{0}\mathbf{1}\{\hat{p}_{in}(\bz_i;k, n) < \frac{\beta\varepsilon\hat{c}_{0}}{(1-\beta)(1-\varepsilon)}\}$, and $\lambda = -\sqrt[m-1]{\frac{(1-\beta)(1 - \varepsilon) k}{\beta\varepsilon c_bn \hat{c}_{0}}}$, we have
$$
\mathbf{1}\{-r_k(\bz_i) \ge \lambda\} = \mathbf{1}\{\hat{p}(g_i = 1|\bz_i) \ge \beta\},
$$
\end{theorem}
where $\hat{p}(\cdot)$ denotes the empirical estimation. The proof is in Appendix~\ref{sup:theory}.

%%%%%%%%%%%%%%%%%%%%%%%%%%%%%%%%%%%%%%%%%%%%%%%%%%%%%%%%%%%%%%
%%%%%%%%%%%%%%%%%%%%%%%  RELATED SECTION %%%%%%%%%%%%%%%%%%%%%
%%%%%%%%%%%%%%%%%%%%%%%%%%%%%%%%%%%%%%%%%%%%%%%%%%%%%%%%%%%%%%

\section{Related Work}
\label{sec:related}

\paragraph{OOD detection}
The phenomenon of neural networks' overconfidence in out-of-distribution data is first revealed in \cite{nguyen2015deep}, which attracts growing research attention in several thriving directions:

\vspace{0.1cm}
(1) One line of work attempted to perform OOD detection by devising scoring functions, including OpenMax score~\citep{openworld}, maximum softmax probability~\citep{Kevin}, ODIN score~\citep{liang2018enhancing}, deep ensembles~\citep{lakshminarayanan2017simple}, Mahalanobis
distance-based score~\citep{lee2018simple}, energy score~\citep{liu2020energy,lin2021mood, wang2021canmulti, morteza2022provable}, activation rectification (ReAct)~\cite{sun2021react},  gradient-based score~\citep{huang2021importance} and ViM score~\cite{wang2022vim}.
In ~\citet{huang2021mos}, the authors revealed that approaches developed for CIFAR datasets might not translate effectively into a large-scale ImageNet benchmark, 
and highlight the need to evaluate OOD detection methods in a real-world setting.
To date, \emph{none} of the prior works investigated the non-parametric nearest neighbor approach for OOD detection. Our work bridges the gap by presenting the first study exploring the efficacy of using nearest neighbor distance for OOD detection.
We demonstrate superior performance on several OOD detection benchmarks, and we hope our work draws attention to the strong promise of the non-parametric approach.

\vspace{0.2cm}
(2) Another promising line of work addressed OOD detection by training-time regularization~\cite{lee2017training, bevandic2018discriminative,  malinin2018predictive, hendrycks2018deep,  geifman2019selectivenet, hein2019relu, meinke2019towards, mohseni2020self, liu2020energy, jeong2020ood, van2020uncertainty, yang2021semantic, chen2021atom, hongxin2022logitnorm, ming2022posterior, katzsamuels2022training}.
For example, models are encouraged to give predictions with uniform distribution~\cite{lee2017training,hendrycks2018deep} or higher energies~\cite{liu2020energy, ming2022posterior, du2022unknown, katzsamuels2022training} for outlier data. Most regularization methods require the availability of auxiliary OOD
data. Recently, VOS~\cite{du2022towards} alleviates the need by automatically synthesizing virtual outliers that can meaningfully regularize the model's decision boundary during training. 

\vspace{0.2cm}
(3) More recently, several works explored the role of representation learning for OOD detection. In particular, \texttt{CSI}~\citep{tack2020csi} investigate the type of data augmentations that are particularly beneficial for OOD detection. Other works~\citep{winkens2020contrastive,2021ssd} verify the effectiveness of applying the off-the-shelf multi-view contrastive losses such as {SimCLR}~\cite{chen2020simple} and {SupCon}~\cite{2020supcon} for OOD detection. These two works both use Mahalanobis distance as the OOD score, and make strong distributional assumptions by modeling the class-conditional feature space as multivariate Gaussian distribution. \citet{ming2022cider} propose a prototype-based contrastive learning framework for OOD detection, which promote stronger ID-OOD separability than SupCon loss. Our method and previous works are fundamentally different in the OOD detection method, despite all benefit from high-quality representations. In particular, \method~is a non-parametric method that does not impose prior of ID distribution. Performance-wise, our method outperforms SSD by a substantial margin, and is easy to use in practice.

\paragraph{KNN for anomaly detection} KNN has been explored for anomaly detection ~\citep{jing2014somknn, zhao2020analysis, liron2020knnanomly}, which aims to detect abnormal input samples from one class. We focus on OOD detection, which {requires additionally performing multi-class classification for ID data}. Some other recent works~\citep{dang2015knntabular, gu2019statknn, pires2020knntabular} explore the effectiveness of KNN-based anomaly detection for the tabular data. The potential of using KNN for OOD detection in deep neural networks is currently underexplored. Our work provides both new empirical insights and theoretical analysis of using the KNN-based approach for OOD detection.

%%%%%%%%%%%%%%%%%%%%%%%%%%%%%%%%%%%%%%%%%%%%%%%%%%%%%%%%%%%%%%
%%%%%%%%%%%%%%%%%%%%%%  CONCLUSION SECTION %%%%%%%%%%%%%%%%%%%
%%%%%%%%%%%%%%%%%%%%%%%%%%%%%%%%%%%%%%%%%%%%%%%%%%%%%%%%%%%%%%
\section{Conclusion}
\label{sec:concl}

This paper presents the first study exploring and demonstrating the efficacy of the non-parametric nearest-neighbor distance for OOD detection. Unlike prior works, the non-parametric approach does not impose {any} distributional assumption about the underlying feature space, hence providing stronger flexibility and generality.
We provide important insights that  
a high-quality feature embedding and a suitable distance measure are two indispensable components for the OOD detection task. Extensive experiments show  KNN-based method can notably improve the performance on several OOD detection benchmarks, establishing superior results. We hope our work inspires future research on using the non-parametric approach to OOD detection.

\section*{Acknowledgement}
Work is supported by a research award from American Family Insurance. Zhu acknowledges NSF grants 1545481, 1704117, 1836978, 2041428, 2023239, ARO MURI W911NF2110317, and MADLab AF CoE FA9550-18-1-0166. The authors would also like to thank ICML reviewers for the helpful suggestions and feedback.

\bibliographystyle{icml2022}
\bibliography{example_paper.bib}

%%%%%%%%%%%%%%%%%%%%%%%%%%%%%%%%%%%%%%%%%%%%%%%%%%%%%%%%%%%%%%%%
%%%%%%%%%%%%%%%%%%%%%%%%%%%%%%%%%%%%%%%%%%%%%%%%%%%%%%%%%%%%%%%%
% APPENDIX
%%%%%%%%%%%%%%%%%%%%%%%%%%%%%%%%%%%%%%%%%%%%%%%%%%%%%%%%%%%%%%%%
%%%%%%%%%%%%%%%%%%%%%%%%%%%%%%%%%%%%%%%%%%%%%%%%%%%%%%%%%%%%%%%%
\newpage
\appendix
\onecolumn

%%%%%%%%%%%%%%%%%%%%%%%%%%%%%%%%%%%%%%%%%%%%%%%%%%%%%%%%%%%%%%
%%%%%%%%%%%%%%%%%%%%%%  THEORY SECTION %%%%%%%%%%%%%%%%%%%%%%%
%%%%%%%%%%%%%%%%%%%%%%%%%%%%%%%%%%%%%%%%%%%%%%%%%%%%%%%%%%%%%%

\section{Theoretical Analysis}
\label{sup:theory}

\paragraph{Proof of Theorem~\ref{th:knn2bayes}} We now provide the proof sketch for readers to understand the key idea, which revolves around performing the empirical estimation of the probability $\hat{p}(g_i = 1|\bz_i)$. By the Bayesian rule, the probability of $\bz$ being ID data is:
\begin{align*}
     p(g_i = 1|\bz_i) &= 
     \frac{p(\bz_i| g_i = 1) \cdot p(g_i = 1)}{p(\bz_i)} \\ 
     &=\frac{p_{in}(\bz_i)\cdot p(g_i = 1)}
     {p_{in}(\bz_i)\cdot p(g_i = 1) + p_{out}(\bz_i)\cdot p(g_i = 0)} \\
    \hat{p}(g_i = 1|\bz_i) &=   \frac{(1-\varepsilon) \hat p_{in}(\bz_i)}{(1-\varepsilon){\hat p_{in}(\bz_i) + \varepsilon\hat p_{out}(\bz_i)}}.
\end{align*}
Hence, estimating $\hat p(g_i = 1|\bz_i)$ boils down to deriving the empirical estimation of $\hat p_{in}(\bz_i)$ and $\hat p_{out}(\bz_i)$, which we show below respectively.

\paragraph{Estimation for $\hat p_{in}(\bz_i)$}

Recall that $\bz$ is a normalized feature vector in $\mathbb{R}^{m}$. Therefore $\bz$ locates on the surface of a $m$-dimensional unit sphere. 
We denote $B(\bz, r) = \{\bz':\lVert\bz'-\bz\rVert_2 \le r\} \cap \{\lVert\bz'\rVert_2 = 1\}$, which is a set of data points on the unit hyper-sphere and are at most $r$ Euclidean distance away from the center $\bz$. Note that the local dimension of $B(\bz, r)$ is $m-1$. 

Assuming the density satisfies Lebesgue's differentiation theorem, the probability density function can be attained by: 
$$ 
p_{in}(\bz_i) = \lim_{r \rightarrow 0} \frac{p(\bz \in B(\bz_i, r)|g_i = 1)}{|B(\bz_i, r)|}.
$$
In training time, we empirically observe $n$ in-distribution samples $\mathbb{Z}_n = \{\bz'_1, \bz'_2, ..., \bz'_n \}$. We assume each sample $\bz'_j$ is \emph{i.i.d} with a probability mass $\frac{1}{n}$. The empirical point-density for the ID data can be estimated by $k$-NN distance:
\begin{align*}
 \hat{p}_{in}(\bz_i;k, n) &= \frac{p(\bz'_j \in B(\bz_i, r_k(\bz_i))|\bz'_j \in \mathbb{Z}_n)}{|B(\bz_i, r_k(\bz_i))|} \\
 &= \frac{k}{c_bn(r_k(\bz_i))^{m-1}}, 
\end{align*}
where $c_b$ is a constant. The following Lemma~\ref{lemma:l1bound} establishes the convergence rate of the estimator.

\begin{lemma}
\label{lemma:l1bound}
$$\lim_{\frac{k}{n} \rightarrow 0} \hat{p}_{in}(\bz_i;k, n) = p_{in}(\bz_i)$$ 
Specifically,
$$\mathbb{E}[|\hat{p}_{in}(\bz_i;k, n) - p_{in}(\bz_i)|] = o(\sqrt[m-1]{\frac{k}{n}}+ \sqrt{\frac{1}{k}})$$
\end{lemma}
The proof is given in~\cite{zhao2020analysis}. 

\paragraph{Estimation for $\hat p_{out}(\bz_i)$}
A key challenge in OOD detection is the lack of knowledge on OOD distribution, which can arise universally outside ID data. We thus try to keep our analysis general and reflect the fact that we do not have any strong prior information about OOD. For this reason, we model OOD data with an equal chance to appear outside of the high-density region of ID data. Our theory is thus complementary to our experiments and captures the universality of OOD data. 
Specifically, we denote $$\hat{p}_{out}(\bz_i) = \hat{c}_{0}\mathbf{1}\{\hat{p}_{in}(\bz_i;k, n) < \frac{\beta\varepsilon\hat{c}_{0}}{(1-\beta)(1-\varepsilon)}\}$$
where the threshold is chosen to satisfy the theorem. 

Lastly, our theorem holds by plugging in the empirical estimation of $\hat p_{in}(\bz_i)$ and $\hat p_{out}(\bz_i)$. 

\begin{proof}

\begin{align*}
    \mathbf{1}\{-r_k(\bz_i) \ge \lambda\} &= \mathbf{1}\{\varepsilon c_b n \hat{c}_{0}(r_k(\bz_i))^{m-1} \le \frac{1-\beta}{\beta}(1-\varepsilon)k\} \\
    &= \mathbf{1}\{\varepsilon c_b n \hat{c}_{0}\mathbf{1}\{\varepsilon c_b n \hat{c}_{0} (r_k(\bz_i))^{m-1} > \frac{1-\beta}{\beta}(1-\varepsilon)k\}(r_k(\bz_i))^{m-1} \le \frac{1-\beta}{\beta}(1-\varepsilon)k\} \\
    &= \mathbf{1}\{\varepsilon c_b n \hat{c}_{0} \mathbf{1}\{\hat{p}_{in}(\bz_i;k, n) < \frac{\beta\varepsilon\hat{c}_{0}}{(1-\beta)(1-\varepsilon)}\}(r_k(\bz_i))^{m-1} \le \frac{1-\beta}{\beta}(1-\varepsilon)k\} \\
    &= \mathbf{1}\{\varepsilon c_b n  \hat{p}_{out}(\bz_i)(r_k(\bz_i))^{m-1} \le \frac{1-\beta}{\beta}(1-\varepsilon)k\} \\
    &= \mathbf{1}\{\frac{k(1 - \varepsilon)}{k(1 - \varepsilon) + \varepsilon c_b n\hat{p}_{out}(\bz_i)(r_k(\bz_i))^{m-1}} \ge \beta\} \\
    &= \mathbf{1}\{\hat{p}(g_i = 1|\bz_i) \ge \beta\}
\end{align*}

\end{proof}

%%%%%%%%%%%%%%%%%%%%%%%%%%%%%%%%%%%%%%%%%%%%%%%%%%%%%%%%%%%%%%
%%%%%%%%%%%%%%%%%%%%%%  THEORY SECTION %%%%%%%%%%%%%%%%%%%%%%%
%%%%%%%%%%%%%%%%%%%%%%%%%%%%%%%%%%%%%%%%%%%%%%%%%%%%%%%%%%%%%%

\begin{table*}[b] 
\centering
\vspace{-0.3cm}
\caption[]{\small \textbf{Comparison results with DenseNet-101.} Comparison with competitive out-of-distribution detection methods. All methods are based on a model trained on {ID data only}.  All values are percentages and are averaged over all OOD test datasets.}
\footnotesize
\begin{tabular}{l|lll|lll}
\toprule
\multirow{2}{*}{\textbf{Method}}  & \multicolumn{3}{c|}{\textbf{CIFAR-10}} & \multicolumn{3}{c}{\textbf{CIFAR-100}} \\
  & \textbf{FPR95} $\downarrow$ & \textbf{AUROC} $\uparrow$ & \textbf{ID ACC} $\uparrow$ & \textbf{FPR95} $\downarrow$ & \textbf{AUROC} $\uparrow$ & \textbf{ID ACC} $\uparrow$ \\ \midrule
MSP & 49.95 & 92.05 & 94.38 & 79.10 & 75.39 & 75.08 \\ 
Energy & 30.16 & 92.44 & 94.38 & 68.03 & 81.40 & 75.08 \\ 
ODIN & 30.02 & 93.86 & 94.38 & 55.96 & 85.16 & 75.08 \\ 
Mahalanobis & 35.88 & 87.56 & 94.38 & 74.57 & 66.03 & 75.08 \\ 
GODIN & 28.98 & 92.48 & 94.22 & 55.38 & 83.76 & 74.50 \\ 
CSI & 70.97 & 78.42 & 93.49 & 79.13 & 60.41 & 68.48 \\ 
SSD+ & 16.21 & 96.96 &  \textbf{94.45} &  43.44 & 88.97  &  \textbf{75.21}\\ 
KNN+ (ours) & \textbf{12.16} & \textbf{97.58} & \textbf{94.45}  & \textbf{37.27} & \textbf{89.63} & \textbf{75.21} \\
  \bottomrule
\end{tabular}
        \vspace{-0.2cm}
        \label{tab:other-arc}
\end{table*}

\vspace{-1cm}
\section{Configurations}
\label{sec:config}

\textbf{Non-parametric methods for anomaly detection} We provide implementation details of the non-parametric methods in this section. Specifically,

\texttt{\textbf{IForest}}~\citep{liu2008iforest} generates a random forest assuming the test anomaly can be isolated in fewer steps. We use 100 base estimators in the ensemble and each estimator draws 256 samples randomly for training. The number of features to train each base estimator is set to 512.   

\texttt{\textbf{LOF}}~\citep{breunig2000lof} defines an outlier score based on the sample’s $k$-NN distances. We set $k=50$. 

\texttt{\textbf{LODA}}~\citep{2016loda} is an ensemble solution combining multiple weaker binary classifiers. The number of bins for the histogram is set to 10. 

\texttt{\textbf{PCA}}~\citep{shyu2003pca} detects anomaly samples with large values when mapping to the directions with small eigenvalues. We use 50 components for calculating the outlier scores. 

\texttt{\textbf{OCSVM}}~\citep{bernhard2001ocsvm} learns a decision
boundary that corresponds to the desired density level set
of with the kernel function. We use the RBF kernel with $\gamma=\frac{1}{512}$. The upper bound on the fraction of training error is set to 0.5. 

Some of these methods~\citep{bernhard2001ocsvm, shyu2003pca} are specifically designed for anomaly detection scenarios that assume ID data is from one class. We show that $k$-NN distance with the class-aware embeddings can achieve both OOD detection and multi-class classification tasks.

\section{Results on Different Architecture}
\label{sec:other_arc}
In the main paper, we have shown that the nearest neighbor approach is competitive on ResNet. In this section, we show in Table~\ref{tab:other-arc} that \method's strong performance holds on different network architectures DenseNet-101~\citep{huang2017densely}. All the numbers reported are averaged over OOD test datasets described in Section~\ref{sec:common_benchmark}. %

\end{document}